\documentclass[letterpaper, 10 pt, conference]{ieeeconf}  % Comment this line out
\IEEEoverridecommandlockouts                              % This command is only
\usepackage[utf8]{inputenc} % Allows UTF-8 input
\usepackage[T1]{fontenc}    % Select font encodings
\usepackage{graphics}       % for pdf, bitmapped graphics files
\usepackage{epsfig}         % for postscript graphics files
\usepackage{times}          % assumes new font selection scheme installed - Use Times font instead
\usepackage{amsmath}        % assumes amsmath package installed
\usepackage{amssymb}        % assumes amsmath package installed
\usepackage{hyperref}       % For hyperlinks, load towards the end

\hypersetup{
    colorlinks=true,
    linkcolor=blue,
    filecolor=magenta,      
    urlcolor=cyan, % Changed email link color for visibility if needed
    pdftitle={Bridging the Generalisation Gap},
    pdfpagemode=FullScreen,
    }
\title{\LARGE \bf
Bridging the Generalisation Gap: Synthetic Data Generation for Multi-Site Clinical Model Validation
}

\author{Bradley Segal$^{1}$, Joshua Fieggen$^{1}$, David A. Clifton$^{1}$, and Lei Clifton$^{1,2}$% <-this % stops a space
\thanks{$^{1}$Computational Health Informatics Lab, Department of Engineering Sciences, University of Oxford, Oxford, UK}%
\thanks{$^{2}$Nuffield Department of Primary Care Health Sciences, University of Oxford, Oxford, UK}%
\thanks{E-mail: \href{mailto:bradley.segal@eng.ox.ac.uk}{bradley.segal@eng.ox.ac.uk}}%
}

\begin{document}

\maketitle
\thispagestyle{empty}
\pagestyle{empty}

%%%%%%%%%%%%%%%%%%%%%%%%%%%%%%%%%%%%%%%%%%%%%%%%%%%%%%%%%%%%%%%%%%%%%%%%%%%%%%%%
\begin{abstract}
Ensuring the generalisability of clinical machine learning (ML) models across diverse healthcare settings remains a significant challenge due to variability in patient demographics, disease prevalence, and institutional practices. Existing model evaluation approaches often rely on real-world datasets, which are limited in availability, embed confounding biases, and lack the flexibility needed for systematic experimentation. Furthermore, while generative models aim for statistical realism, they often lack transparency and explicit control over factors driving distributional shifts. In this work, we propose a novel structured synthetic data framework designed for the \textit{controlled benchmarking} of model robustness, fairness, and generalisability. Unlike approaches focused solely on mimicking observed data, our framework provides \textit{explicit control} over the data generating process, including site-specific prevalence variations, hierarchical subgroup effects, and structured feature interactions. This enables targeted investigation into \textit{how} models respond to specific distributional shifts and potential biases. Through controlled experiments, we demonstrate the framework's ability to isolate the impact of site variations, support fairness-aware audits, and reveal generalisation failures, particularly highlighting how model complexity interacts with site-specific effects. This work contributes a reproducible, interpretable, and configurable tool designed to advance the reliable deployment of ML in clinical settings.
\end{abstract}

%%%%%%%%%%%%%%%%%%%%%%%%%%%%%%%%%%%%%%%%%%%%%%%%%%%%%%%%%%%%%%%%%%%%%%%%%%%%%%%%
\section{INTRODUCTION}
\label{sec:introduction}

Machine learning (ML) models for healthcare must generalise across diverse clinical settings, yet real-world deployment often reveals substantial degradation in performance due to shifts in patient demographics, disease prevalence, and site-specific practice patterns~\cite{yangMachineLearningGeneralizability2022, barak-correnPredictionHealthcareSettings2021, sunMachineLearningBased2022}. Such challenges, vividly illustrated by failures in transporting models for tasks like COVID-19 screening between hospitals~\cite{yangMachineLearningGeneralizability2022} or predicting emergency room dispositions across different facilities~\cite{barak-correnPredictionHealthcareSettings2021}, are exacerbated in multi-site clinical studies. Addressing these issues requires robust validation strategies that systematically account for distributional shifts and site heterogeneity.

Traditional model validation relies on real-world multi-site datasets. However, real data access is often limited by privacy restrictions, data-sharing limitations, cost, inherent biases~\cite{ahmadEnhancingDataAccessibility2024}, and crucially, they hinder systematic isolation of failure modes under varying conditions. Existing synthetic data approaches, such as generative adversarial networks (GANs) and variational autoencoders (VAEs), aim to replicate the statistical properties of real-world data~\cite{haoSyntheticDataAI2024}. However, these methods often lack explicit control over distributional factors like site-specific prevalence or structured feature interactions crucial for rigorously evaluating model generalisability and fairness mitigation strategies~\cite{goyalSystematicReviewSynthetic2024}. There is a need for tools that bridge this gap, offering control alongside plausible data generation.

To address these challenges, we propose a structured synthetic data framework designed to enable explicit control over key factors influencing model generalisability, including site-specific variations, hierarchical subgroup effects, and feature interactions. Unlike generative approaches that prioritise statistical realism often at the cost of transparency, our method allows for targeted benchmarking of model generalisability, fairness, and robustness in simulated clinical settings. Our framework provides configurability over: multi-site data generation with user-defined prevalence variations (e.g., simulating differing disease burden across hospitals); hierarchical subgroup effects based on demographic factors (e.g., modelling age-sex specific risk profiles); complex feature interactions, including non-linear dependencies; temporal dependencies and structured missing data patterns; and explicit modelling of feature-outcome relationships, providing a known ground truth for validation.

By enabling structured control over these elements, the framework allows researchers to systematically study model performance under specific, challenging conditions relevant to real-world deployment. While this structured approach prioritises interpretability and control over potentially replicating every unknown statistical nuance present in real data, this focus is crucial for systematically studying the mechanisms underlying generalisation failures. This emphasis on transparency makes the framework particularly useful for benchmarking models, evaluating domain adaptation techniques, and conducting fairness-aware audits. A detailed comparison with alternative approaches is provided in Section~\ref{sec:related_work}.

Our contributions in this work include:
\begin{itemize}
    \item A configurable framework for generating multi-site clinical data with explicit control over site-specific variations, subgroup effects, and feature interactions relevant to generalisability.
    \item Systematic evaluation demonstrating how controlled variations in site-specific interactions impact model generalisability, enabling the assessment of external validation performance under specific distributional shifts.
    \item Analysis illustrating the framework's utility in examining the relationship between model complexity and generalisability across sites.
    \item An open-source implementation facilitating reproducible benchmarking of clinical ML models under controlled, challenging conditions.
\end{itemize}
We validate our framework through controlled experiments that assess its ability to preserve feature-outcome relationships, capture realistic site-level variations, and provide actionable insights into model robustness.

The remainder of this paper is organised as follows: Section~\ref{sec:related_work} discusses related work, Section~\ref{sec:methods} describes the framework’s technical architecture, Section~\ref{sec:results} presents validation experiments, and Section~\ref{sec:discussion} discusses broader applications, limitations, and future extensions.

%%%%%%%%%%%%%%%%%%%%%%%%%%%%%%%%%%%%%%%%%%%%%%%%%%%%%%%%%%%%%%%%%%%%%%%%%%%%%%%%
\section{Related Work}
\label{sec:related_work}

The challenge of generating realistic and useful synthetic clinical data has spurred research across several distinct approaches. Our work builds upon and differentiates itself from these existing methodologies, particularly concerning the objective of evaluating model generalisability across diverse clinical settings.

\subsection{Generative Models}
Deep generative models (GANs~\cite{goodfellowGenerativeAdversarialNetworks2020}, VAEs~\cite{kingmaAutoEncodingVariationalBayes2022}) can achieve high statistical fidelity by learning complex, high-dimensional distributions found in real clinical data~\cite{yoonAnonymizationDataSynthesis2020, yoonTimeseriesGenerativeAdversarial}. Their strength lies in automatically capturing intricate correlations without explicit feature engineering. However, they typically lack mechanisms for explicit, fine-grained control over specific factors critical for generalisability testing, such as site-specific outcome prevalence, interaction strengths, or the precise nature of subgroup effects~\cite{goyalSystematicReviewSynthetic2024}. Systematically varying one factor or simulating specific counterfactuals can be difficult, hindering targeted simulation of distributional shifts relevant to multi-site validation. Our framework contrasts this by prioritising explicit, configurable control over data-generating mechanisms.

\subsection{EHR Simulation Tools}
Tools like Synthea~\cite{walonoskiSyntheaApproachMethod2018} focus on simulating realistic Electronic Health Record (EHR) structures and longitudinal patient journeys based on health statistics and clinical guidelines. While valuable for tasks requiring realistic EHR workflows (e.g., testing data pipelines), they are generally not designed for benchmarking ML model generalisability under controlled statistical variations. Their primary focus is plausible patient pathways, not providing the fine-grained control over distributions, feature-outcome relationships, site effects, or interaction strengths needed to systematically probe ML model robustness to known challenges like prevalence shifts or demographic bias. Our framework complements these tools by focusing on datasets tailored for ML validation with known ground truth.

\subsection{Real Multi-Site Data Validation}
Validating models on large, real-world multi-site datasets~\cite{ahmadEnhancingDataAccessibility2024} is the gold standard but faces significant practical hurdles: patient privacy concerns (e.g., GDPR, HIPAA), data sharing agreements, curation/harmonisation costs, and ethical approvals~\cite{ahmadEnhancingDataAccessibility2024}. Furthermore, real data inherently contains confounding factors and unknown biases, making it difficult to isolate specific reasons for performance degradation or perform controlled experiments varying only one factor (e.g., site prevalence). Our framework provides a complementary approach: a controlled, accessible, reproducible environment for systematic pre-validation and benchmarking.

\subsection{Procedural and Rule-Based Generation}
Simpler procedural or rule-based methods are fully interpretable but often struggle to capture the complex correlations, non-linearities, and interaction effects present in real clinical data. Our framework builds on this transparency but incorporates more sophisticated effect modelling, including interactions, non-linearities, and hierarchical effects, offering greater flexibility and realism for simulating complex clinical scenarios relevant to ML validation.

In summary, our framework occupies a unique position by blending the transparency of procedural generation with the capability to model complex, multi-level effects, specifically targeting the need for controlled benchmarking of ML model generalisability in heterogeneous clinical environments.

%%%%%%%%%%%%%%%%%%%%%%%%%%%%%%%%%%%%%%%%%%%%%%%%%%%%%%%%%%%%%%%%%%%%%%%%%%%%%%%%
\section{Methods}
\label{sec:methods}

\subsection{Data Generation Framework}
\label{subsec:data_gen_framework}

Our framework uses a configurable pipeline comprising three core modules (feature generation, effect modelling, and outcome generation) to create synthetic multi-site clinical data with explicit ground truth relationships, balancing realism with control. Unlike traditional approaches replicating statistical properties post hoc, our framework inverts the process. It begins by constructing a population-level risk model based on user-specified variable characteristics. Model coefficients (effect strengths) are sampled (e.g., from normal distributions) based on predefined effect sizes for main effects and interaction terms (complexity governed by configurable maximum order and sampling probability). Feature values are then generated by sampling from chosen distributions (e.g., normal, uniform, categorical) to match configured characteristics. Site-specific prevalence targets (via \texttt{site\_prevalence\_range} or \texttt{target\_average\_prevalence}) are achieved by adjusting internal decision thresholds.

Outcomes are generated by applying the modelled risk to the feature data and adjusting decision thresholds to match site-specific prevalence targets. Controlled stochastic variation around these thresholds mimics real-world diagnostic uncertainty. This design assumes underlying risk factors drive outcomes, but observed prevalence varies contextually. Decoupling relationship modelling from feature sampling allows efficient generation of datasets with arbitrary sample sizes while maintaining consistent underlying relationships. A fixed random seed ensures reproducibility.

The risk model at a specific time point is formalised as:
\begin{equation}
\label{eq:risk_model}
\text{Logit}(\text{Risk}_i) = \eta_i = \sum_{j=1}^{n} \beta_j X_{ij} + \sum_{k=1}^{m} \gamma_k Z_{ik} + \epsilon_i
\end{equation}
where $X_{ij}$ represents predictive features for individual $i$ weighted by $\beta_j$, $Z_{ik}$ represents interaction terms with effect sizes $\gamma_k$, and $\epsilon_i$ models random noise. Risk probability is obtained via a link function, e.g., $\text{Risk}_i = \text{sigmoid}(\eta_i)$. For longitudinal data, temporal dependencies are introduced during feature generation (Section~\ref{subsubsec:feature_gen}).

The framework is initialised with a configuration object defining key parameters, including: data structure parameters (number of samples, sites, target prevalence ranges/means per site); feature characteristics (number and types of predictive/noise features); effect modelling parameters (interaction complexity, non-linear transformations, effect scaling); missing data rates; and demographic distributions (population-level distributions, subgroup effect sizes).

\subsubsection{Feature Generation}
\label{subsubsec:feature_gen}
The feature generation module creates predictive features (correlated with the outcome) and noise features (uncorrelated, although can be correlated to other noise features). Continuous features are sampled (e.g., Normal, Uniform), and categorical features are generated with variable category counts and distributions.

For longitudinal data, structured temporal variation is introduced. Feature values can evolve via specified dynamics (e.g., autoregressive processes, transition matrices) simulating trends or treatment effects while maintaining within-patient consistency. Custom feature generators can be incorporated.

\subsubsection{Effect Modelling}
\label{subsubsec:effect_model}
Effect modelling integrates multiple variability sources into the ground truth relationship: main feature effects (linear associations); demographic subgroup-specific effects (variations in baseline risk or feature effects across groups); site-specific variations and interactions (adjustments based on clinical site); feature interactions (higher-order, configurable); and non-linear transformations (e.g., quadratic, log, exponential) applied to features.

To ensure interpretability and stability, effect strengths are dynamically scaled. Interaction terms, especially higher-order ones, can otherwise dominate the risk score, overshadowing weaker effects. To mitigate this artificial amplification, interaction effect sizes are scaled inversely with interaction order $p$. The default scaling for an order-$p$ interaction involving features $X_1, \dots, X_p$ is:
\begin{equation}
\label{eq:scaling}
\gamma_{p} = \frac{\gamma}{\sqrt{p}}
\end{equation}
where $\gamma$ is the base interaction effect size and $\gamma_p$ the scaled coefficient. This maintains balanced contributions but can be adjusted or disabled.

\subsection{Site and Subgroup Effects}
\label{subsec:site_subgroup}

The framework incorporates site-specific variations to simulate heterogeneous settings. Site-level differences are modelled through several mechanisms: site-specific baseline effects (configurable intercept shifts); site-specific feature interactions (configurable modifications to $\beta_j$ depending on site); prevalence matching (site-specific decision thresholds on risk scores enforce prevalence targets); and stochastic adjustments (controlled noise prevents rigid decision boundaries). This explicit modelling allows targeted testing of model robustness and domain adaptation under known shifts. For example, one could simulate differing biomarker predictive value across sites by configuring site-specific interactions.

In addition to site heterogeneity, demographic subgroup effects are modelled hierarchically: main subgroup effects (baseline risk adjustments by group); feature-specific subgroup modifications (adjustments to $\beta_j$ based on subgroup); and a hierarchical structure where site variations can be nested within broader demographic trends. This structure allows simulation of conditions with strong demographic associations (e.g., cardiovascular risk influenced by age/sex) or scenarios where dynamic physiological features dominate risk over baseline demographics (e.g., ICU sepsis prediction).

%%%%%%%%%%%%%%%%%%%%%%%%%%%%%%%%%%%%%%%%%%%%%%%%%%%%%%%%%%%%%%%%%%%%%%%%%%%%%%%%
\section{Results}
\label{sec:results}

\subsection{Framework Validation}

To evaluate the performance and fidelity of the proposed data generation framework, we conducted a series of validation experiments across varying configurations. The results demonstrate the framework’s ability to generate synthetic datasets with controllable characteristics, reliably preserve specified statistical relationships, and scale efficiently for typical clinical modelling tasks.

\subsubsection{Data Generation Performance}
We assessed the framework's computational efficiency across dataset configurations typical of clinical studies. Table~\ref{tab:configurations} summarises the runtime and memory usage for generating datasets with different numbers of sites, features (predictive and noise), and samples. The framework demonstrates predictable scaling, with generation times remaining under 6 seconds for moderately sized configurations (up to 10 sites, 30 features, 10,000 samples), indicating suitability for rapid prototyping and experimentation.

% TABLE I: Computational Requirements for Standard Clinical Configurations
\begin{table*}[h]
\caption{Computational Requirements for Standard Configurations}
\label{tab:configurations}
\begin{center}
\begin{tabular}{|c|c|c|c|c|c|}
\hline
Configuration & Sites & Features & Samples & Runtime (s) & Memory (MB) \\
\hline
Basic   & 3  & 10  & 1000  & 3.71  & 153.7 \\
Medium  & 5  & 20  & 5000  & 4.21  & 156.9 \\
Large   & 10 & 30  & 10000 & 5.49  & 158.7 \\
\hline
\end{tabular}
\end{center}
\footnotesize{Runtime and memory usage (mean over 5 runs) for generating datasets with varying numbers of sites, features (predictive and noise), and samples using a standard configuration (interaction order 3). Tests performed on an AMD Ryzen 7 5800X 8-core CPU with 32 GB of RAM.}
\end{table*}

To further probe scalability under more demanding conditions, we increased feature counts, sample sizes, and the maximum interaction order allowed during effect modelling. Table~\ref{tab:scalability} presents performance metrics across these more complex configurations. As expected, runtime increases substantially with both the number of features and the interaction order, primarily due to the combinatorial growth in potential feature interactions considered during effect generation. The transition from `High' (60 features, order 4, 10k samples) to `Very High' (60 features, order 5, 100k samples) highlights this, with runtime increasing significantly. Memory usage scales more directly with the number of samples.

% TABLE II: Performance Impact of Increasing Simulation Complexity
\begin{table*}[h]
\caption{Performance Impact of Increasing Simulation Complexity}
\label{tab:scalability}
\begin{center}
\begin{tabular}{|c|c|c|c|c|c|c|}
\hline
Complexity & Sites & Features & Samples & Interaction Order & Runtime (s) & Memory (MB) \\
\hline
Low    & 3  & 14 & 1000   & 2 & 4.02 $\pm$ 0.09 & 160.9 $\pm$ 0.1 \\
Medium & 5  & 30 & 5000   & 3 & 4.27 $\pm$ 0.01 & 163.9 $\pm$ 0.1 \\
High   & 10 & 60 & 10000  & 4 & 14.65 $\pm$ 0.67 & 181.2 $\pm$ 0.6 \\
Very High & 20 & 60 & 100000 & 5 & 243.48 $\pm$ 10.74 & 325.6 $\pm$ 1.9 \\
\hline
\end{tabular}
\end{center}
\footnotesize{Runtime and memory usage (mean $\pm$ standard deviation over 5 runs) for generating datasets with increasing complexity (sites, features, samples, max interaction order). Tests performed on an AMD Ryzen 7 5800X 8-core CPU with 32 GB of RAM.}
\end{table*}

These results suggest that while the framework efficiently handles configurations common in clinical research, users should anticipate increased computational cost when simulating scenarios with very high feature dimensionality or complex, high-order interactions. The Discussion section provides further context on managing computational complexity.

\subsubsection{Effect Recovery Analysis}
\begin{figure}[thpb]
    \centering
    \includegraphics[width=0.95\columnwidth]{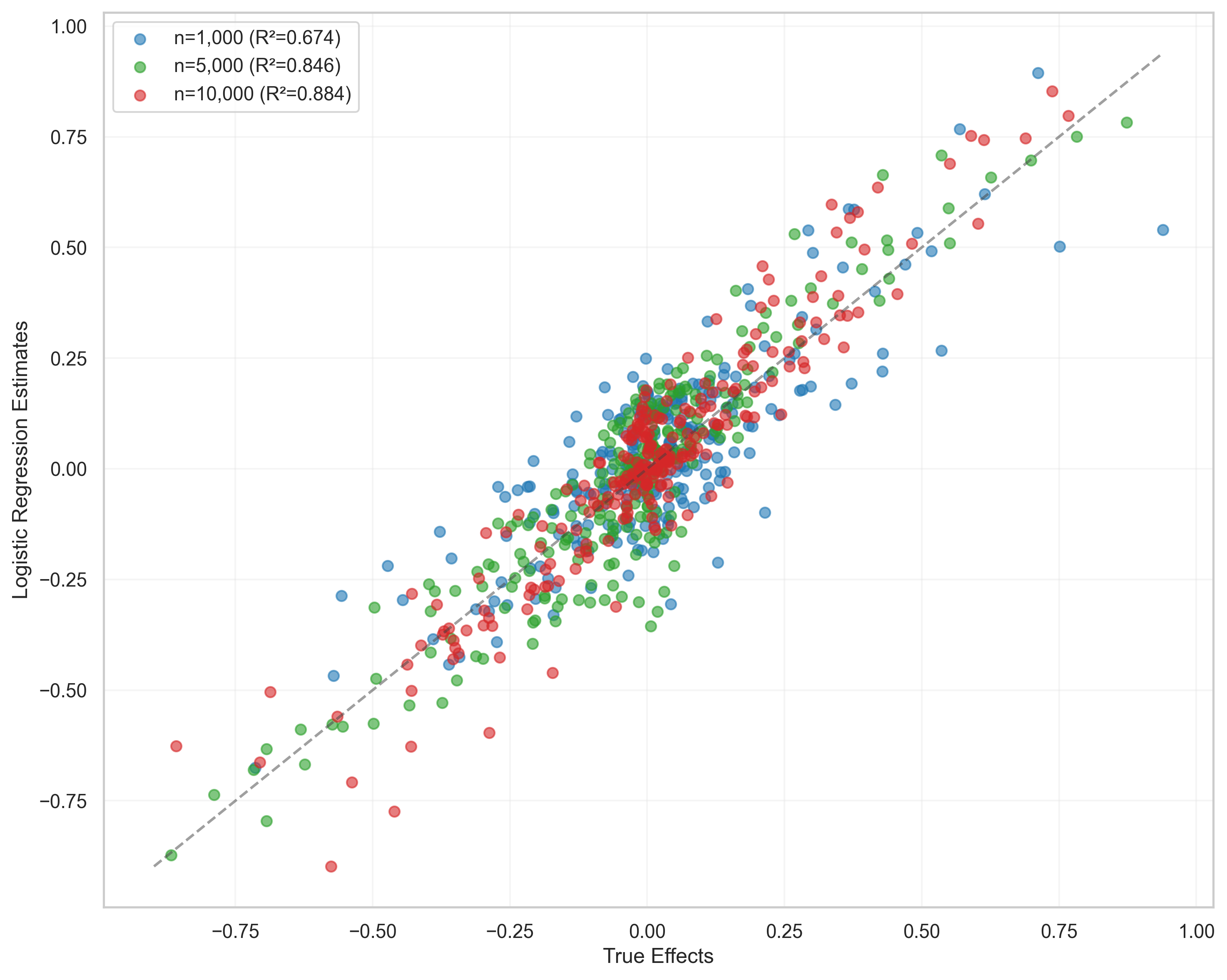} 
    \caption{Comparison of specified (true) and recovered (observed) feature effect sizes (logits) from logistic regression models trained on generated datasets of varying sizes (N=1k, 5k, 10k). Points represent individual features; alignment along the diagonal indicates accurate effect recovery.}
    \label{fig:effect_recovery}
\end{figure}

A crucial aspect of the framework is its ability to embed known ground truth relationships that can be recovered through analysis. We assessed this fidelity by comparing the specified feature effect coefficients (used internally by the framework's risk model) with coefficients recovered by training external logistic regression models on the generated datasets. Figure~\ref{fig:effect_recovery} plots the true (specified) logit effect size against the recovered logit effect size for individual features across multiple dataset sizes. The strong alignment along the diagonal indicates accurate preservation and recovery of the intended feature effects.

% FIGURE 1: Effect Recovery Scatter Plot

\begin{figure}[thpb]
    \centering
    \includegraphics[width=0.95\columnwidth]{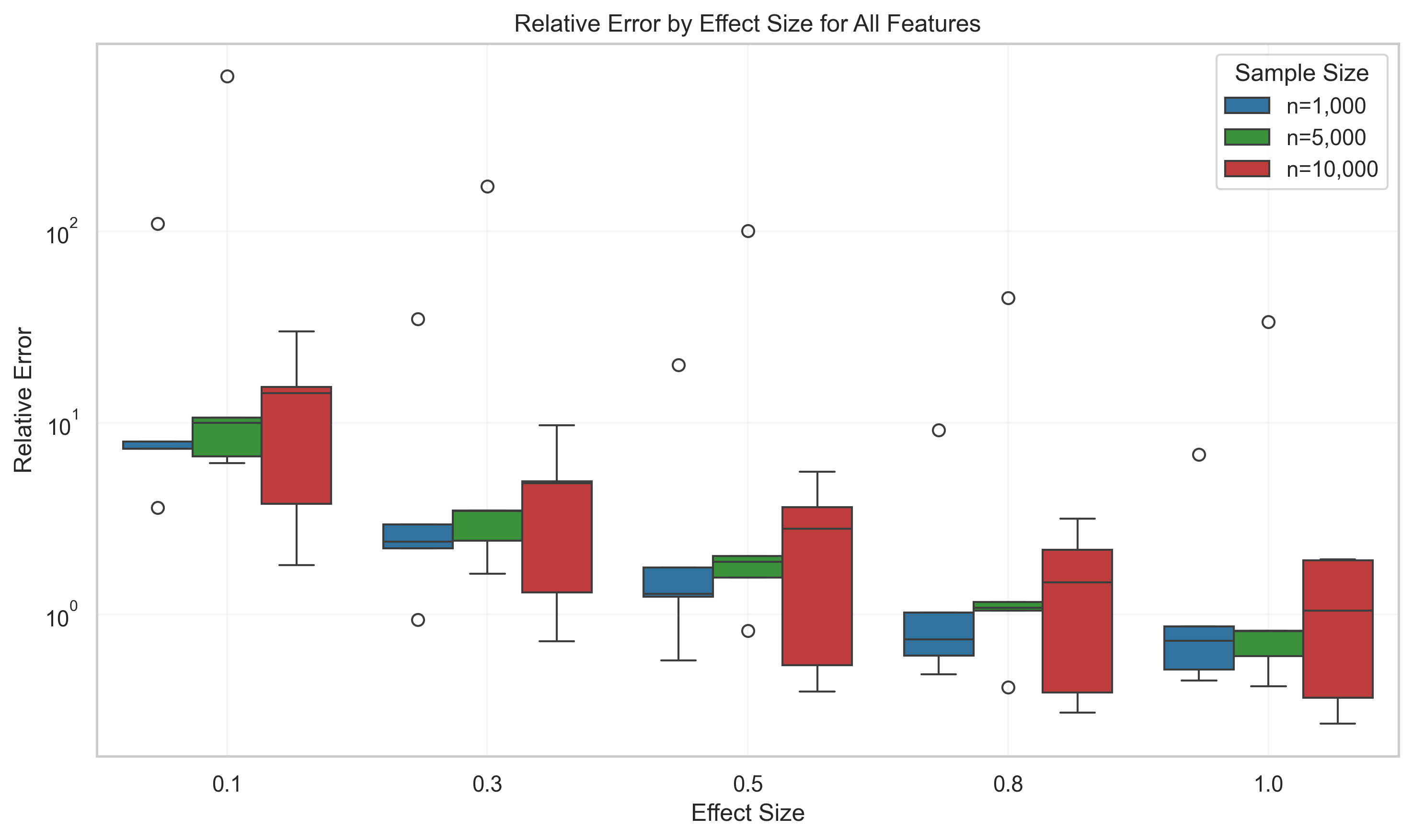} 
    \caption{Relative recovery error of feature effect sizes as a function of the true effect magnitude for different dataset sizes (N). Error diminishes as both sample size and effect magnitude increase, indicating improved recovery precision.}
    \label{fig:effect_recovery_size}
\end{figure}

Figure~\ref{fig:effect_recovery_size} examines the relative recovery error (difference between true and recovered effect size, normalised by true effect size) as a function of the true effect magnitude for different sample sizes. Consistent with statistical theory, the recovery error decreases as both the sample size and the true effect magnitude increase. Larger effects are recovered with higher precision, and larger datasets reduce sampling variability, leading to more accurate estimates even for weaker effects. This confirms the framework's ability to reliably embed ground truth relationships suitable for model validation and benchmarking.

% FIGURE 2: Relative Error Plot

\subsubsection{Site Variation Control}
\begin{figure}[thpb]
    \centering
    \includegraphics[width=0.95\columnwidth]{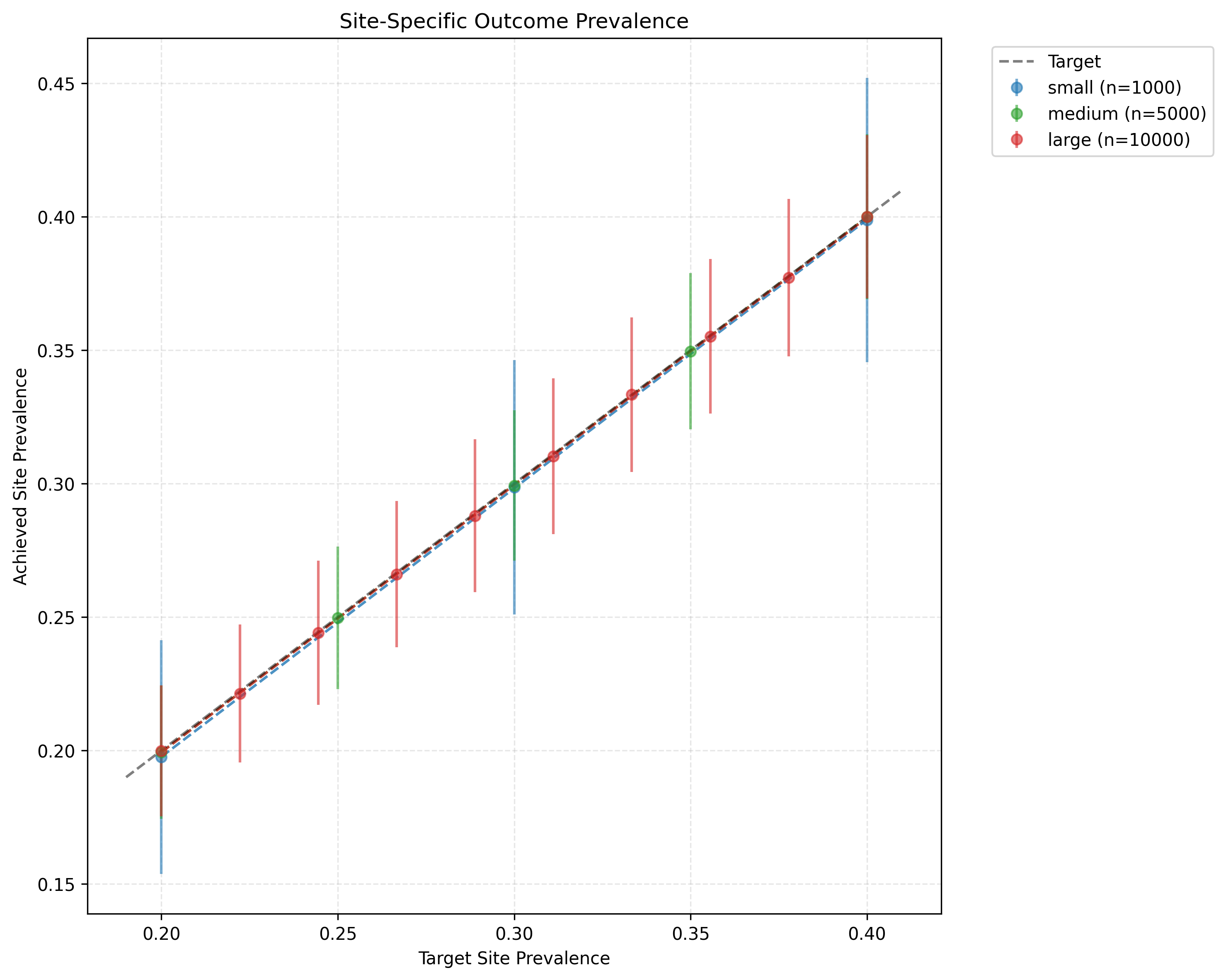} % Ensure path is correct
    \caption{Observed vs. Target Outcome Prevalence per Site. Each point represents a simulated clinical site. Error bars show 95\% bootstrapped confidence intervals for the observed prevalence within generated datasets of different sizes (N=1k, 5k, 10k). Observed prevalences closely match target values, with decreasing variability at larger sample sizes.}
    \label{fig:site_variation}
\end{figure}

The framework's ability to control site-specific outcome prevalence is critical for simulating heterogeneous clinical environments. We evaluated this by comparing the target prevalence specified for each site during configuration with the observed prevalence in the generated datasets across different sample sizes. Figure~\ref{fig:site_variation} presents these results, showing minimal deviation between target and observed prevalence rates (mean absolute difference $<$ 0.01\%). As expected, smaller sample sizes exhibit greater variability due to random sampling effects, but the central tendencies closely match the specified targets. Bootstrap resampling was used to estimate the confidence intervals for observed prevalence within each generated dataset.

% FIGURE 3: Site Variation Plot

These findings confirm the framework's effectiveness in generating multi-site datasets with precisely controlled, site-specific outcome prevalence variations, a key capability for evaluating model generalisability and calibration across different clinical settings.

\subsection{Predictive Model Performance}
To demonstrate the utility of the generated data for downstream ML tasks, we assessed the performance and stability of standard classification algorithms trained on a synthetic dataset. The dataset comprised 5000 samples, 10 continuous and 10 categorical predictive features, and 2 noise features, simulating a typical clinical prediction task. We trained Logistic Regression (LR), Random Forest (RF)~\cite{breimanRandomForests2001}, Gradient Boosting (GB)~\cite{friedmanGreedyFunctionApproximation2001}, and XGBoost (XGB)~\cite{chenXGBoostScalableTree2016}, evaluating performance using standard metrics averaged over 5 runs with different random seeds.

% TABLE III: Predictive Model Performance
\begin{table*}[h]
\caption{Predictive Model Performance on Generated Data}
\label{tab:model_performance}
\begin{center}
% Note: Removed ↑ arrows as they are non-standard table elements
\begin{tabular}{|l|c|c|c|c|c|c|}
\hline
Model & AUROC & Sens. & Spec. & PPV & NPV & F1 \\
\hline
LR  & 0.833 (0.803-0.863) & 0.691 (0.648-0.870) & 0.817 (0.633-0.867) & 0.622 (0.501-0.685) & 0.858 (0.841-0.920) & 0.655 (0.617-0.704) \\
RF  & 0.850 (0.824-0.878) & 0.794 (0.647-0.849) & 0.758 (0.707-0.893) & 0.588 (0.545-0.723) & 0.894 (0.849-0.919) & 0.676 (0.638-0.726) \\
GB  & 0.856 (0.828-0.882) & 0.700 (0.639-0.815) & \textbf{0.858 (0.754-0.901)} & \textbf{0.683 (0.585-0.753)} & 0.868 (0.842-0.903) & 0.691 (0.651-0.739) \\
XGB & \textbf{0.884 (0.860-0.907)} & \textbf{0.807 (0.733-0.864)} & 0.819 (0.773-0.869) & 0.660 (0.605-0.729) & \textbf{0.907 (0.877-0.934)} & \textbf{0.726 (0.686-0.768)} \\
\hline
\end{tabular}
\end{center}
\footnotesize{Performance metrics (mean and 95\% confidence interval across 5 trials) for classification models trained on a standard synthetic dataset (N=5000, 10 continuous/10 categorical predictive features). Area Under the ROC Curve (AUROC), Sensitivity (Sens.), Specificity (Spec.), Positive Predictive Value (PPV), Negative Predictive Value (NPV), F1-score, Logistic Regression (LR), Random Forest (RF), Gradient Boosting (GB), XGBoost (XGB).}
\end{table*}

Table~\ref{tab:model_performance} shows that the generated data supports effective model training, with performance patterns consistent with typical observations on real clinical data~\cite{subudhiComparingMachineLearning2021}. XGBoost achieved the highest overall performance, followed by Gradient Boosting and Random Forest, with Logistic Regression serving as a linear baseline. The relatively tight confidence intervals suggest stable performance across runs. This confirms that the framework generates data with sufficient complexity and signal to differentiate between ML algorithms effectively.

\subsection{Model Generalisability Analysis}
\label{subsec:gen_analysis}

To specifically demonstrate the framework's utility in assessing model generalisability, we designed an experiment simulating deployment across different clinical sites with varying site-specific dynamics. We focused on evaluating how models handle changing feature relationships across sites, a key challenge for transportability. This was achieved by systematically varying the strength of site-feature interaction effects, which model scenarios where a feature's predictive contribution differs depending on the clinical site.

We generated datasets across multiple configurations where the interaction effect size (defined here as the standard deviation of the normal distribution from which site-feature interaction coefficients were sampled, prior to scaling) was varied from 0.0 (no site-specific interactions) to 1.5. A fixed interaction probability of 0.5 was maintained, ensuring that roughly half the site-feature pairs had non-zero interactions, allowing us to isolate the impact of interaction magnitude. Each dataset included 8 simulated clinical sites, with data from 20\% of sites held out as an external validation set. Models were trained on the remaining 80\% of sites using internal cross-validation. We measured performance degradation, defined as the absolute difference in Area Under the ROC Curve (AUROC) between the internal validation estimate and the performance on the external validation sites, averaged over 5 trials.

% FIGURE 4: Generalisability Plot
\begin{figure}[thpb]
    \centering
    \includegraphics[width=0.95\columnwidth]{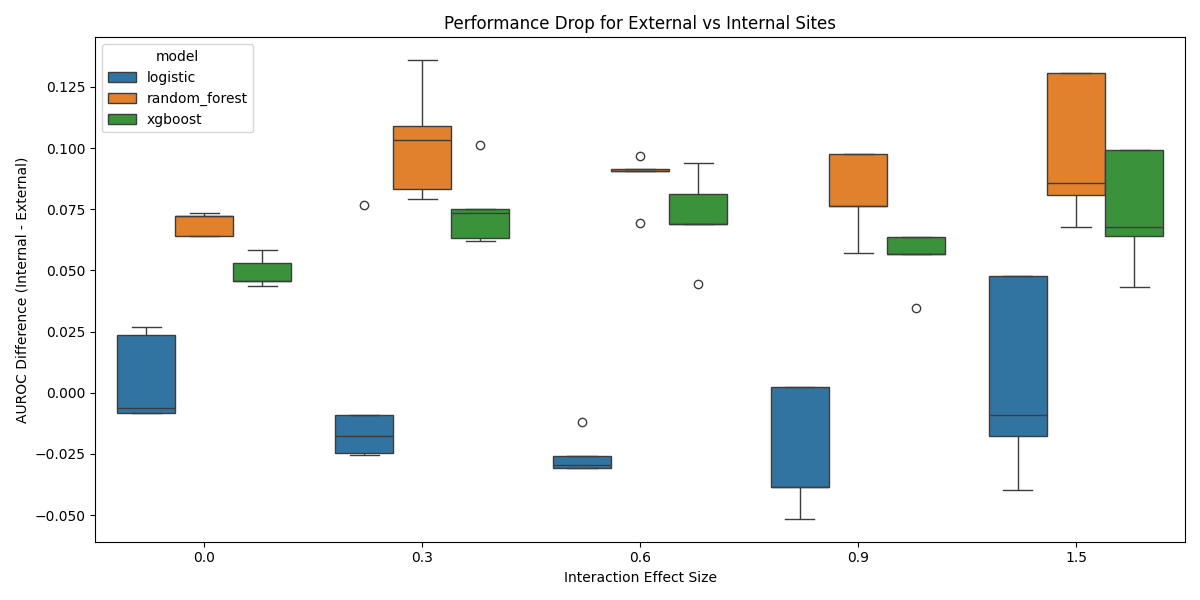} 
    \caption{Model Performance Degradation on External Sites with Increasing Site-Feature Interaction Effects. Performance degradation is the drop in AUROC between internal cross-validation and external site validation. Higher values indicate poorer generalisability. Models: Logistic Regression (LR), Random Forest (RF), Gradient Boosting (GB), XGBoost (XGB).}
    \label{fig:external_site_performance}
\end{figure}

The results, shown in Figure~\ref{fig:external_site_performance}, reveal distinct generalisability patterns based on model complexity. While the more complex models (RF, GB, XGB) achieved higher absolute AUROC on internal validation (Table~\ref{tab:model_performance}), they exhibited greater performance degradation on the external sites as the site-feature interaction effect size increased. The linear model (LR) showed more stable performance across sites, with minimal degradation regardless of interaction strength. This suggests that the complex models potentially overfit to the site-specific interaction patterns present in the training sites, a vulnerability not fully captured by standard internal cross-validation. The performance gap widened approximately linearly up to an interaction effect size of roughly 0.6, beyond which the degradation seemed to plateau, possibly indicating a limit to how much these specific interactions influenced the outcome relative to main effects.

These findings underscore the utility of this framework for simulating challenging multi-site deployment scenarios. It allows researchers to quantify how specific types of distributional shifts (here, site-feature interactions) impact models of varying complexity, potentially revealing generalisation issues that might be missed using internal validation alone. This supports more informed model selection and highlights the need for robust evaluation strategies when deploying models across heterogeneous environments.

%%%%%%%%%%%%%%%%%%%%%%%%%%%%%%%%%%%%%%%%%%%%%%%%%%%%%%%%%%%%%%%%%%%%%%%%%%%%%%%%
\section{Discussion}
\label{sec:discussion}

The persistent challenge of ensuring clinical prediction models generalise reliably across diverse healthcare settings, highlighted by studies from Yang et al.~\cite{yangMachineLearningGeneralizability2022}, Barak-Corren et al.~\cite{barak-correnPredictionHealthcareSettings2021}, and Sun et al.~\cite{sunMachineLearningBased2022}, underscores a critical barrier to the widespread, trustworthy implementation of clinical ML. Factors such as site-specific processes, varying patient demographics, and differing outcome prevalence contribute to performance degradation when models are deployed outside their training environment~\cite{laskoWhyProbabilisticClinical2024}. Prospectively identifying these failure points is difficult due to the limited availability of large, accessible multi-site datasets and the inherent complexity of real-world clinical data.

Existing validation paradigms relying solely on retrospective real-world data~\cite{ahmadEnhancingDataAccessibility2024} face practical limitations including data access restrictions, privacy concerns, and the inability to disentangle confounding factors. While deep generative models like GANs and VAEs offer alternatives by learning complex data distributions~\cite{haoSyntheticDataAI2024}, they often lack the transparency and explicit control needed to systematically probe specific generalisability mechanisms. Our proposed framework addresses this gap by providing a simulation environment that prioritises control and interpretability over perfect statistical mimicry. Its core value lies in enabling researchers to systematically investigate how specific factors, such as site-prevalence shifts or the strength of subgroup interactions, impact model performance under known ground truth conditions. This controlled approach facilitates the understanding of failure modes, which is often obscured in purely data-driven approaches. For instance, a researcher could use the framework to simulate the scenario described by Lasko et al.~\cite{laskoWhyProbabilisticClinical2024} where a model fails when transported between sites with different underlying prevalence rates, isolating prevalence shift as the failure cause.

The framework's utility extends to rigorously assessing model robustness and facilitating fairness audits. By allowing users to explicitly configure demographic subgroup effects and site-specific feature interactions, it serves as a tool for studying potential biases, rather than being a source of unintended bias itself. Researchers can simulate scenarios where specific subgroups might experience differential model performance (e.g., due to feature effects varying by age or sex) and evaluate the effectiveness of bias mitigation strategies in a controlled setting~\cite{obermeyerDissectingRacialBias2019}. This proactive approach to fairness assessment is crucial for developing equitable clinical decision support systems.

One of the primary advantages of this framework is its ability to disentangle the effects of site heterogeneity from other sources of variation. As demonstrated in our generalisability experiments (Section~\ref{subsec:gen_analysis}), systematically modulating site-feature interaction strengths allows for a granular examination of how different model architectures respond to specific distributional shifts. This capability is particularly valuable for evaluating domain adaptation methods and selecting models with appropriate complexity for multi-site deployment, potentially identifying overfitting risks that standard cross-validation might miss.

While offering extensive configurability, the framework has inherent limitations. The trade-off for explicit control is that the generated data may not capture all unknown complex correlations or rare event patterns present in real-world data. The current implementation of longitudinal dynamics, while capturing basic temporal dependencies, could be enhanced to model more complex disease trajectories or treatment responses. Furthermore, computational demands increase significantly with higher feature dimensionality and interaction orders, as shown in Table~\ref{tab:scalability}. Users should be mindful of this combinatorial complexity, starting with simpler configurations and judiciously increasing complexity based on the specific research question and available resources.

Future work will focus on bridging the gap between controlled simulation and real-world complexity. Hybrid modelling approaches, perhaps using real-world marginal distributions to parameterise the framework's feature generators or using the framework's output to condition or fine-tune deep generative models, could enhance statistical realism while retaining interpretability. Incorporating more sophisticated structured missingness mechanisms, such as simulating missing not at random (MNAR) scenarios where missingness depends on outcomes or other features, would improve the framework's utility for evaluating imputation techniques and model robustness to incomplete data.

To ensure reproducibility and encourage adoption, this framework is available at \href{https://github.com/BradSegal/ClinicalDataSimulator}{Clinical Multi-Site Data Simulator} as an open-source package.

%%%%%%%%%%%%%%%%%%%%%%%%%%%%%%%%%%%%%%%%%%%%%%%%%%%%%%%%%%%%%%%%%%%%%%%%%%%%%%%%
\section{Conclusion}
\label{sec:conclusion}

The increasing adoption of machine learning in clinical settings necessitates rigorous evaluation methodologies to ensure model reliability, fairness, and generalisability across diverse healthcare environments. This work presented a novel structured synthetic data generation framework designed specifically to facilitate the systematic assessment of predictive models under controlled conditions that mimic multi-site deployment challenges. By incorporating explicit, configurable control over site-specific variations, hierarchical subgroup effects, and structured feature interactions, the framework enables targeted experimentation on model transportability and robustness against known sources of distributional shift.

Our results demonstrate that structured synthetic data provides an effective alternative to solely relying on real-world datasets for benchmarking and stress-testing clinical machine learning models. The framework offers a tool for researchers seeking to evaluate model performance under different assumptions, understand failure modes related to data heterogeneity, detect potential biases across demographic subgroups, and develop strategies for mitigating distributional shifts. By leveraging its configurability and known ground truth, the approach can serve as a foundation for advancing research in fairness-aware learning, domain adaptation techniques, and the validation of federated machine learning systems.

Future work will focus on enhancing the realism of synthetic data distributions, potentially through hybrid generative approaches, and by incorporating more sophisticated methods for modelling structured missing data. Expanding the framework's capabilities to accommodate larger-scale simulations and systematic comparisons with real-world validation studies will further solidify its role in clinical AI research. Ultimately, by providing a scalable, transparent, and interpretable approach to model evaluation, this framework contributes to the crucial ongoing effort to deploy trustworthy and effective machine learning systems in healthcare.

%%%%%%%%%%%%%%%%%%%%%%%%%%%%%%%%%%%%%%%%%%%%%%%%%%%%%%%%%%%%%%%%%%%%%%%%%%%%%%%%
\section*{Acknowledgments}

BS and JF are supported by Rhodes Scholarships. DAC is supported by the Pandemic Sciences Institute at the University of Oxford; the National Institute for Health Research (NIHR) Oxford Biomedical Research Centre (BRC); an NIHR Research Professorship; a Royal Academy of Engineering Research Chair; the Wellcome Trust funded VITAL project (Grant 204904/Z/16/Z); the EPSRC (Grant EP/W031744/1); and the InnoHK Hong Kong Centre for Cerebro-cardiovascular Engineering (COCHE). The ADH group at the Nuffield Department of Primary Care Health Sciences is supported by the National Institute for Health and Care Research (NIHR) Applied Research Collaboration Oxford and Thames Valley at Oxford Health NHS Foundation Trust. The views expressed are those of the author(s) and not necessarily those of the NHS, the NIHR or the Department of Health and Social Care.

% Balance columns on last page
\addtolength{\textheight}{-5.5cm}   % Adjust this value as needed

%%%%%%%%%%%%%%%%%%%%%%%%%%%%%%%%%%%%%%%%%%%%%%%%%%%%%%%%%%%%%%%%%%%%%%%%%%%%%%%%
% References are important section.
% Make sure you have a references.bib file with all your citations.
\bibliography{references}

%%%%%%%%%%%%%%%%%%%%%%%%%%%%%%%%%%%%%%%%%%%%%%%%%%%%%%%%%%%%%%%%%%%%%%%%%%%%%%%%
% Appendixes should appear before the acknowledgment if needed.
% \section*{APPENDIX}
% Appendix content goes here.
%%%%%%%%%%%%%%%%%%%%%%%%%%%%%%%%%%%%%%%%%%%%%%%%%%%%%%%%%%%%%%%%%%%%%%%%%%%%%%%%

\end{document}